\documentclass{article}

\usepackage[preprint]{neurips_2026}

\usepackage[utf8]{inputenc}
\usepackage[T1]{fontenc}
\usepackage{graphicx}
\usepackage{booktabs}
\usepackage{amsmath}
\usepackage{amssymb}
\usepackage{hyperref}
\usepackage{caption}
\usepackage{enumitem}
\usepackage{float}
\usepackage{microtype}
\usepackage{xcolor}
\usepackage{placeins}

\graphicspath{{paperGraphs_post_refusal/}}

\newcommand{\changedN}[1]{#1}
\newcommand{\changedV}[1]{#1}
\newcommand{\changedW}[1]{#1}
\newenvironment{changedWblock}{}{}

\title{ConfidenceBench: Evaluating Confidence Calibration\\
  in Large Language Models}

\author{%
  \begin{tabular}[t]{cc}
    Matthew ffrench-Constant & Daniel Yang \\
    Independent Researcher   & ETH Zurich \\
    \texttt{matt.ffrench-constant@cantab.ac.uk} & \texttt{daniel.yang@inf.ethz.ch} \\[1em]
    Xinmeng Huang & Sanyam Kapoor \\
    University of Pennsylvania & New York University \\
    \texttt{xinmengh@sas.upenn.edu} & \texttt{sanyam@nyu.edu} \\
  \end{tabular}
}

\begin{document}

\maketitle

\begin{abstract}
Large language models (LLMs) are increasingly deployed in settings where fluent but
incorrect answers can be costly.
In these settings, accuracy alone is insufficient: models must also know when they
are likely to be wrong.
We present \textbf{ConfidenceBench}, a calibration benchmark that evaluates verbalized
confidence estimates in 15 frontier LLMs using the Brier score, a proper scoring rule
that incentivises truthful probability reporting.
Confidence is elicited via prompting, requiring no access to model logits and making
the framework applicable to both closed-source and open-source systems.
The benchmark comprises 200 private multiple-choice questions across four categories:
spatial reasoning, high-precision mathematics, word lookup, and unknowable questions.
Across three independent runs, Claude Opus 4.6 and Gemini 3.1 Pro Preview achieve
the lowest Brier scores, both reported as $0.103$.
Both substantially outperform the calibrated-random baseline of $0.1875$, while
Gemini 3.1 Flash-Lite scores $0.367$, indicating severe miscalibration.
Accuracy and calibration diverge substantially across model families: the most
accurate model is not the best-calibrated, and several models perform worse than
the calibrated-random Brier baseline despite reasonable accuracy.
These results show that verbalized confidence calibration is a distinct and
practically important axis of LLM reliability, complementary to standard
accuracy-based evaluation.
\end{abstract}

\section{Introduction}

Large language models (LLMs) are increasingly deployed in settings where incorrect
answers can be costly, from software engineering and scientific assistance to
education, medicine, and decision support \changedV{\citep{corea2026}}.
Yet even highly capable models frequently produce fluent but incorrect outputs,
hallucinate unsupported claims \changedV{\citep{kadavath2022language}}, or answer
questions that should be treated as uncertain.
In such settings, accuracy alone is insufficient: a model that is wrong with low
confidence may still be useful, while a model that is wrong with high confidence
can be actively harmful.

Reliable uncertainty estimates are therefore a prerequisite for safe and effective
deployment.
If an LLM can accurately report when it is likely to be wrong, downstream systems
can route difficult queries to stronger models, abstain from answering, request
human review, or allocate additional computation only when needed.
This makes confidence calibration central not only to model evaluation, but also
to practical deployment decisions in reliability-critical applications
\citep{corea2026}.

Uncertainty quantification (UQ) provides a principled framework for this problem.
For LLMs, two broad families of UQ methods exist.
\emph{White-box} methods extract uncertainty from internal token-level probability
distributions, providing theoretically grounded estimates but requiring direct
access to model internals, which is often unavailable for commercial frontier models
served via API.
\emph{Black-box verbalized} methods instead prompt models to state their confidence
explicitly, yielding a universally applicable signal at the cost of relying on the
model's self-knowledge \citep{kadavath2022language,lin2022uncertainty,tian2023calibration}.
Verbalized elicitation is especially attractive because it can be applied uniformly
across closed-source and open-source systems without requiring logits, model weights,
or repeated sampling.

Despite growing interest in LLM calibration, existing benchmarks remain largely
accuracy-centric.
Standard evaluations such as MMLU \citep{hendrycks2020mmlu} and BIG-Bench
\citep{srivastava2022bigbench} tell us what models answer correctly, but not whether
their stated confidence is reliable.
Recent uncertainty benchmarks have begun to address this gap, but many aggregate
broad datasets or focus on abstention \changedV{\citep{ubench2025,ye2024benchmarking,kirichenko2025abstention}}
rather than directly evaluating verbalized probability calibration across frontier
models with a proper scoring rule.
As a result, we still lack a compact benchmark designed specifically to isolate
different calibration failure modes in modern LLMs.

We introduce \textbf{ConfidenceBench}, a calibration benchmark for evaluating
verbalized confidence in frontier LLMs.
The benchmark comprises 200 private multiple-choice questions across four categories:
spatial reasoning, high-precision mathematics, word lookup, and unknowable questions.
These categories are designed to expose qualitatively distinct epistemic failure
modes, including overconfidence on plausible but false recall, multi-step reasoning
errors, and uncertainty under genuinely inaccessible information.

Our contributions are threefold.
First, we present a controlled private benchmark that directly evaluates stated
confidence rather than accuracy alone.
Second, we evaluate 15 frontier LLMs across three independent runs using the Brier
score, a proper scoring rule that incentivises truthful probability reporting.
Third, we provide a human baseline and category-level analyses showing that accuracy
and calibration can diverge substantially across model families.
Together, these results demonstrate that confidence calibration is an essential
dimension of LLM reliability, complementary to standard accuracy-based evaluation.

\section{Related Work}

\changedN{\paragraph{Confidence elicitation.}
A central question in LLM reliability research is whether models can accurately
assess and report their own uncertainty.
Early work by \citet{kadavath2022language} and \citet{lin2022uncertainty} shows
that models can express calibrated confidence when prompted appropriately.
Subsequent studies demonstrate that simple elicitation strategies (asking models
to report probabilities) can yield meaningful confidence estimates even in
black-box settings \citep{tian2023calibration,xiong2024can}\changedV{\citep{yang2024verbalized}}.

This has become increasingly important as access to token-level probabilities is
often restricted in proprietary APIs.
While logit-based methods provide theoretically grounded uncertainty estimates,
they are impractical in many real-world deployments.
Verbalized confidence therefore offers a model-agnostic alternative, though it
requires careful evaluation due to systematic overconfidence.
Notably, reinforcement learning from human feedback
(RLHF\changedV{; \citealt{christiano2017rlhf,ouyang2022training}}) has been shown
to distort internal probability estimates \citep{kapoor2024taught}, further
motivating black-box elicitation approaches based on prompting rather than logits.

\paragraph{Uncertainty estimation methods.}
Beyond verbalized confidence, a broad literature explores alternative UQ methods,
including logit-based confidence \citep{guo2017calibration}, entropy measures
\changedV{\citep{fadeeva2023lm}}, and sampling-based approaches such as
self-consistency \changedV{\citep{xiong2024can}}.
Recent surveys \citep{liu2025survey} categorise these methods into internal,
sampling-based, and linguistic approaches, each with trade-offs in reliability and
computational cost.
Sampling-based approaches can improve uncertainty estimation by measuring variability
across outputs, but require multiple forward passes.
In contrast, verbalized confidence provides a single-pass, scalable signal suitable
for API-based evaluation.
ConfidenceBench focuses on this practical, universally applicable setting.

\paragraph{Calibration benchmarks.}
Several recent benchmarks explicitly evaluate LLM uncertainty.
UBench \citep{ubench2025} aggregates multiple existing datasets to provide broad
coverage, while \citet{ye2024benchmarking} demonstrate that incorporating
uncertainty significantly alters model rankings.
\changedV{\citet{yang2024verbalized} provide a dedicated benchmark of verbalized
confidence scores, showing that prompt design substantially affects calibration
quality across models.}
Other frameworks covered in recent surveys \citep{liu2025survey} provide broad
evaluation tooling but emphasise coverage over targeted isolation of uncertainty
regimes.

Recent work has also begun to systematise black-box and benchmark-based evaluation
of LLM uncertainty.
\citet{xiong2024can} study confidence elicitation in black-box LLMs, covering
prompting strategies, sampling methods, aggregation techniques, calibration, and
failure prediction.
They find that verbalized confidence is often overconfident, although calibration
tends to improve with model capability.
LM-Polygraph \citep{fadeeva2023lm} provides an engineering framework for
uncertainty estimation in text generation, implementing multiple methods and an
extensible benchmark interface.
MAQA \citep{yang2025maqa} studies uncertainty under data ambiguity by constructing
questions with multiple valid answers, highlighting that uncertainty estimation can
degrade when the answer space is plural rather than single-valued.

A key line of work evaluates model awareness of unanswerable questions.
\citet{yin2023selfaware} measure whether models recognise when they lack knowledge,
and AbstentionBench \citep{kirichenko2025abstention} evaluates abstention behaviour
across diverse unanswerable scenarios, finding that reasoning fine-tuning can
degrade abstention performance and lead to confident hallucinations.
Domain-specific benchmarks \citep{testoni2026clinical} further show that calibration
varies significantly across domains, reinforcing the need for targeted evaluation.
ConfidenceBench differs from prior work by evaluating a compact, controlled set
of uncertainty regimes under a unified verbalized-confidence and Brier-score
framework, including an Unknowable category based on empirically verifiable but
inaccessible facts.

\paragraph{Evaluation metrics.}
Calibration is commonly measured using Expected Calibration Error
\citep[ECE;][]{naeini2015bbq}, but this metric is sensitive to binning choices
\citep{kumar2019verified}.
Proper scoring rules provide a more principled alternative: the Brier score
\citep{brier1950verification,gneiting2007strictly} evaluates squared error between
predicted probabilities and outcomes, and is uniquely minimised when predictions
reflect true beliefs.
Recent work increasingly favours the Brier score for calibration evaluation as it
avoids binning artefacts; comprehensive UQ evaluation frameworks \citep{lafage2025torch}
similarly employ it alongside complementary metrics.
Formal definitions of all metrics used in this paper appear in
Section~\ref{sec:metrics}.

ConfidenceBench complements this literature through four design choices.
First, it uses explicit verbalized probability elicitation, enabling black-box
evaluation across frontier models.
Second, it uses a compact, private, manually authored question set designed to
isolate distinct uncertainty regimes, including a controlled Unknowable category.
Third, it evaluates 15 frontier models across repeated runs with run-to-run
stability reporting.
Fourth, it uses the Brier score as the primary metric, directly rewarding
calibrated probability reporting.
Unlike broad uncertainty toolkits or aggregate benchmarks, ConfidenceBench is
intended as a targeted stress test of confidence calibration rather than a
comprehensive measure of general capability.}

\section{Benchmark Design and Methodology}
\label{sec:design}

\subsection{Question Set}

The benchmark comprises 200 multiple-choice questions (50 per category), kept
private to prevent training data contamination.
The four categories are:

\begin{enumerate}[leftmargin=*,topsep=2pt,itemsep=1pt]
\item \textbf{Spatial Reasoning}: dynamic physical scenarios requiring careful
mental simulation of multiple interacting objects or steps.
\textit{Example:} A marble is placed into a mug with a hole in the
bottom, which is then flipped upside down onto a table. Where is the marble?

\item \textbf{High-Precision Mathematics}: multi-step calculations where a
single rounding error or conceptual slip produces a plausible but wrong answer.
\textit{Example:} Take the third decimal digit of $\sqrt{867}$,
multiply by $\sqrt{456}$, and round to the nearest integer.

\item \textbf{Word Lookup}: specific word recall from well-known texts; wrong
options are all plausible alternatives, so the answer cannot be guessed.
\textit{Example:} What is the fifth word of Chapter Seven of
\textit{Harry Potter and the Chamber of Secrets}?

\item \textbf{Unknowable}: questions whose correct answers cannot be determined
without direct prior access to specific real-world objects or events.
\textit{Example:} What was the colour of the kiosk on the corner of
Santa Engracia and R\'ios Rosas, Madrid, on 1st April 2026?
\end{enumerate}

All 200 questions were manually authored and verified, drawn from
no existing dataset or benchmark.
\changedN{Questions were authored by the first author without LLM assistance,
eliminating any risk of overlap between the benchmark and model training data.
The four categories isolate distinct epistemic failure modes rather than testing
domain knowledge breadth: overconfidence on known unknowns (Unknowable),
surface-level over-guessing (Word Lookup), multi-step error accumulation
(Mathematics), and mental simulation gaps (Spatial).}
Correct answers were independently confirmed: mathematics answers via direct
computation; spatial, word-lookup, and unknowable answers cross-checked against
primary sources.
Wrong options are plausible alternatives that require genuine knowledge or
computation to reject, not process of elimination.

\subsection{Confidence Elicitation}

Each question is appended with a prompt requesting a JSON response: \texttt{answer}
(A--D) and \texttt{probability} (0--100, $100 =$ certain, $25 =$
calibrated-random baseline for a four-choice question).
The full prompt template is reproduced in Appendix~\ref{app:prompt}.
Unparseable responses and explicit refusals are scored as $25\%$
confidence with a randomly selected answer rather than being excluded.
Exclusion would inflate apparent calibration for models that refuse most often
on difficult questions, systematically biasing their scores upward.

\subsection{Evaluation Metrics}
\label{sec:metrics}

\changedN{Let $p_i \in [0,1]$ be the stated probability (elicited on a 0--100 scale
and divided by 100) and $y_i \in \{0,1\}$ the correctness indicator.

\paragraph{Brier score.}
\begin{equation}
  \bar{B} = \frac{1}{n}\sum_{i=1}^{n}(p_i - y_i)^2
  \label{eq:brier}
\end{equation}
Lower values indicate better calibration.
For a four-choice question answered with uniformly random guessing at $p=0.25$,
the expected Brier score is $0.1875$, which serves as our calibrated-random
baseline.
The Brier score is a \emph{proper scoring rule} \citep{gneiting2007strictly}:
it is uniquely minimised in expectation when a model reports its true belief,
directly aligning incentives with honest uncertainty reporting.

\paragraph{Calibration gap.}
\begin{equation}
  G = \frac{1}{n}\sum_{i=1}^{n}(p_i - y_i)
  \label{eq:calgap}
\end{equation}
The only difference from the Brier score is the absence of the square, making
overconfidence ($G > 0$) and underconfidence ($G < 0$) directly visible as
signed quantities.

\paragraph{Expected Calibration Error.}
\begin{equation}
  \text{ECE} = \sum_{b=1}^{B}\frac{|B_b|}{n}\,\bigl|\bar{p}_b - \bar{y}_b\bigr|
  \label{eq:ece}
\end{equation}
where $B_b$ is the set of samples in bin $b$ \citep{naeini2015bbq}.
Unlike the Brier score, ECE requires choosing a bin count $B$; we use $B=10$
equal-width bins, noting that results are sensitive to this choice
\citep{kumar2019verified}.
We also report kernel-smoothed reliability diagrams
(smECE; \citealt{blasiok2023smoothece}) in Appendix~\ref{app:reliability}
because smECE avoids discontinuities introduced by bin boundaries while remaining
interpretable as a calibration summary.}

\section{Experimental Setup}

We evaluate 15 frontier models (Table~\ref{tab:models}).
GPT-5 High, Standard, and Low share the same base model (\texttt{gpt-5}) and
differ only in \path{reasoning_effort}, enabling a controlled within-family
comparison.
GPT-5 Mini and Nano are distinct, smaller base models that extend the capability
range.
Each model runs all 200 questions three times at temperature~1.0.
We report the mean and standard deviation of each metric across the three runs.

\paragraph{Human baseline.}
An external volunteer completed all 200 questions in a single session with internet
access, with exactly one minute per question.
The tester used the same A--D answer format and provided a numeric confidence
estimate for each response, making the human baseline directly comparable to model
outputs under identical elicitation conditions.
Results are reported alongside model scores throughout the Results section.

\changedN{\paragraph{Compute.}
Evaluation was conducted exclusively via commercial APIs.
Each model received 600 total queries (200 questions $\times$ 3 runs), for a
total of approximately 9{,}000 API calls across the 15 models evaluated.}

\begin{table}[t]
\vspace{0.5\baselineskip}
\centering
\small
\caption{Models evaluated in ConfidenceBench.}
\label{tab:models}
\vspace{0.5\baselineskip}
\begin{tabular}{ll}
\toprule
\textbf{Model} & \textbf{Provider} \\
\midrule
GPT-5 High / Standard / Low  & OpenAI    \\
GPT-5 Mini / Nano             & OpenAI    \\
\midrule
Claude Opus 4.6 / 4.5 / 4.1  & Anthropic \\
Claude Sonnet 4.6 / 4.5 / 4  & Anthropic \\
\midrule
Gemini 3.1 Pro Preview        & Google    \\
Gemini 3.1 Flash-Lite         & Google    \\
Gemini 2.5 Pro / Flash        & Google    \\
\bottomrule
\end{tabular}
\end{table}

\section{Results}

\subsection{Calibration and Accuracy}

\begin{figure}[t]
  \centering
  \includegraphics[width=\textwidth]{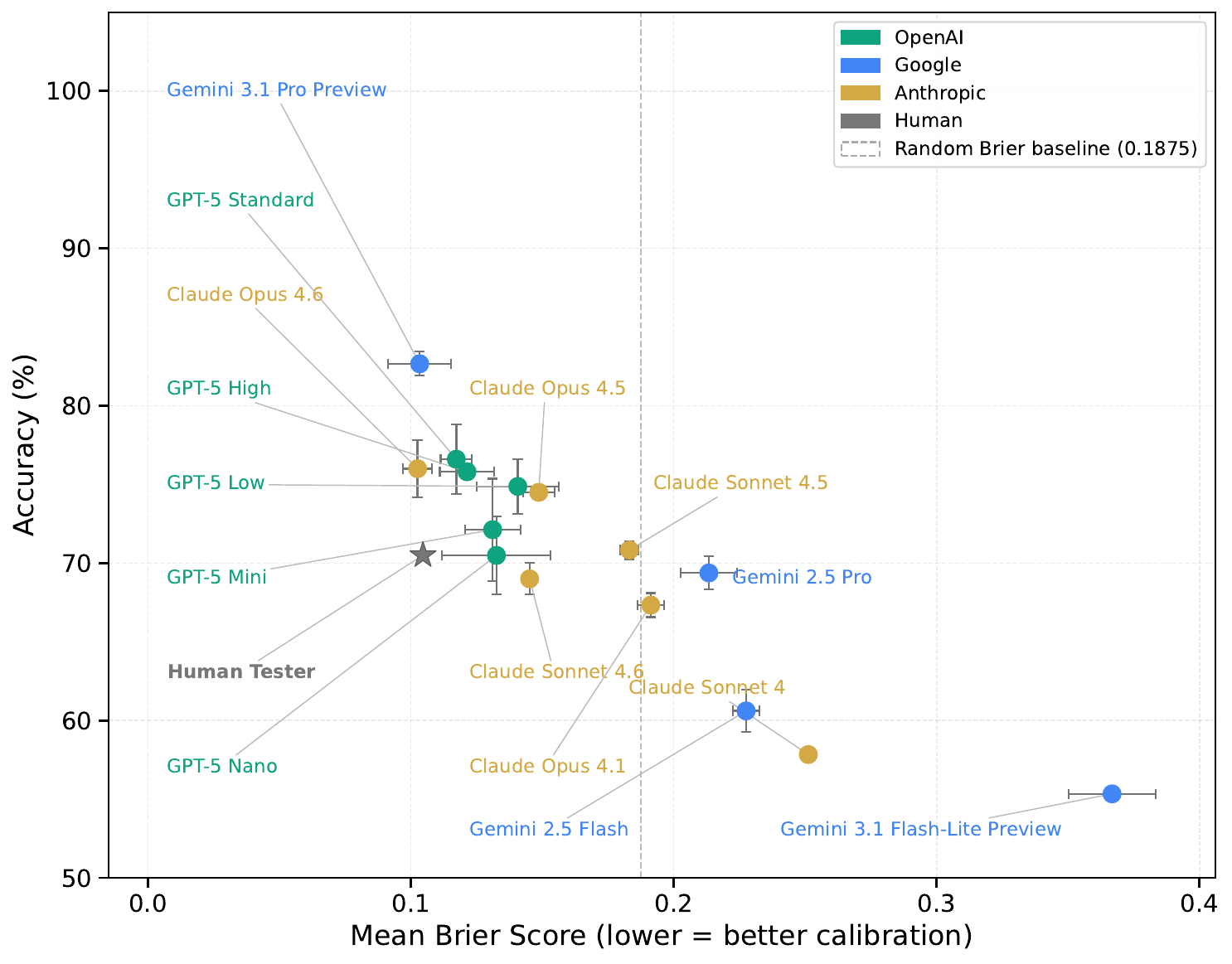}
  \caption{Brier score vs.\ accuracy (3-run means). Error bars show standard
  deviation in both dimensions (horizontal: Brier score, vertical: accuracy).
  Upper-left is optimal: low Brier, high accuracy. Dashed lines show
  calibrated-random baselines. The most accurate model is not the best-calibrated.}
  \label{fig:scatter}
\end{figure}

Figure~\ref{fig:scatter} plots each model in Brier-vs-accuracy space.
The most accurate model (Gemini 3.1 Pro Preview, $82.7\%$) is not the
best-calibrated (Claude Opus 4.6, Brier $0.103$), confirming that calibration
and accuracy measure qualitatively different aspects of model reliability.
Mean Brier scores per model are shown in Figure~\ref{fig:brier_scores}
(Appendix~\ref{app:additional_figs}).
A cluster of two models and the human tester occupy the top tier (Brier
$0.102$--$0.105$), while five of fifteen models perform worse than the
calibrated-random baseline of $0.1875$.
The human tester achieved overall accuracy $70.5\%$ and Brier $0.105$, placing
within the top three models.
Accuracy rankings are also shown in Appendix~\ref{app:additional_figs}.

\subsection{Results by Question Type}

\begin{figure}[t]
  \centering
  \includegraphics[width=\columnwidth]{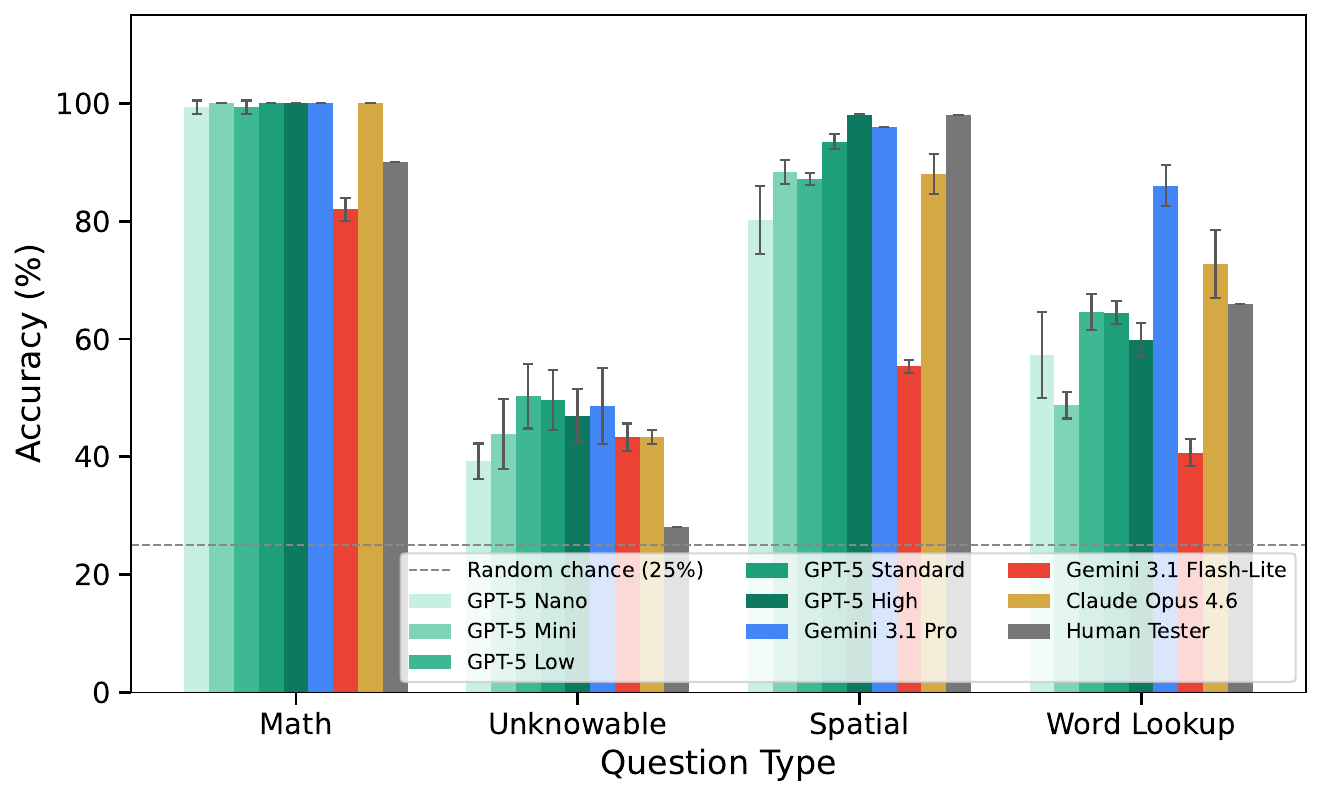}
  \caption{Accuracy by question type for selected models and the human baseline
  (3-run means). GPT-5 family ordered by reasoning effort; human shown for
  reference.}
  \label{fig:accuracy_by_type}
\end{figure}

Figure~\ref{fig:accuracy_by_type} shows per-category accuracy for selected models
and the human baseline.
Mathematics is straightforward for nearly all models, with high accuracy and
well-calibrated confidence.
Conversely, Gemini 3.1 Flash-Lite performs poorly on
mathematics in both accuracy and calibration.
Unknowable questions are the dominant miscalibration driver: models correctly
recognise they are guessing but default to $25\%$ (the ``I don't know'' floor)
even when they empirically answer correctly far more often.
The human tester achieved per-category accuracy of $88\%$ (mathematics), $92\%$
(spatial), $56\%$ (word lookup), and $28\%$ (unknowable).
Brier scores by question type are provided in Appendix~\ref{app:additional_figs}.

\begin{figure}[htbp]
  \centering
  \includegraphics[width=\columnwidth]{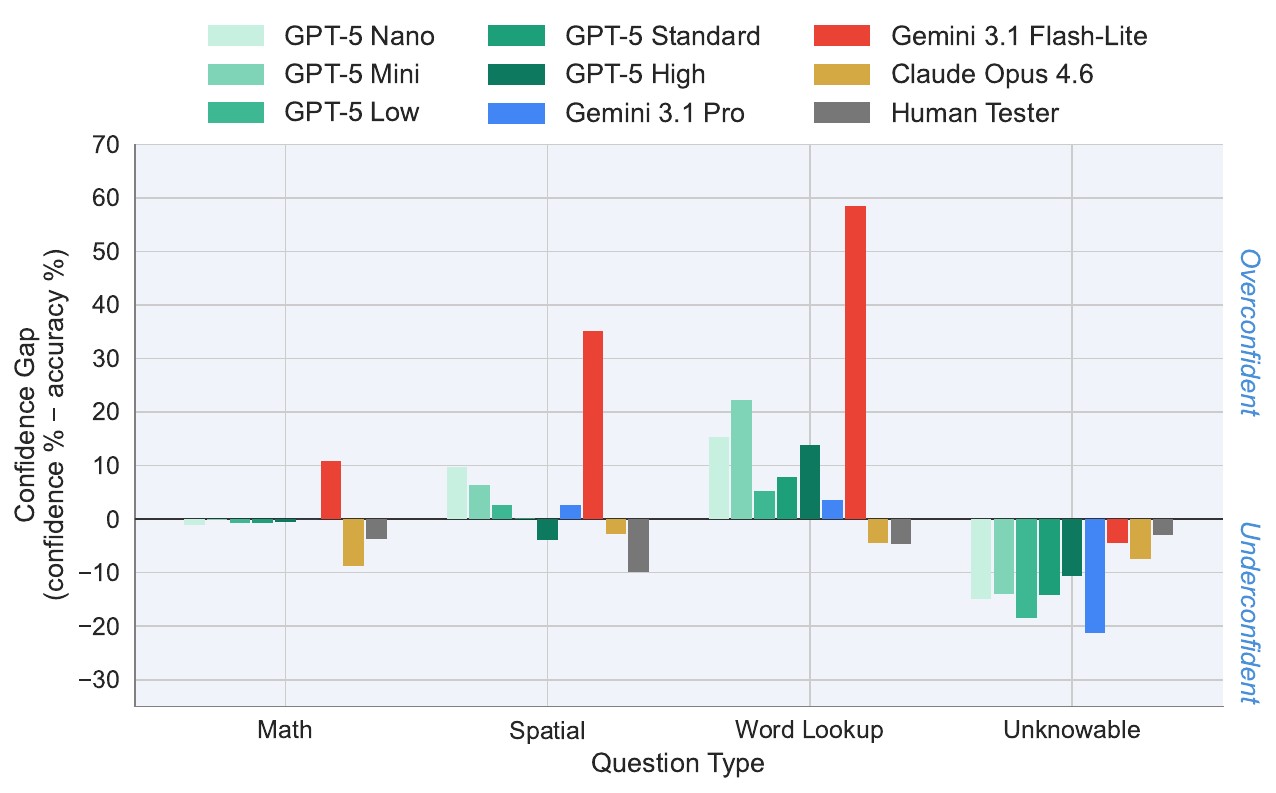}
  \caption{Calibration gap (mean stated confidence minus actual accuracy) by
  question type for selected models.
  Negative = underconfident; positive = overconfident; zero = perfectly calibrated.
  3 runs pooled.
  \changedW{Claude Opus 4.6 (gold) shows consistent underconfidence across all
  four question types.}}
  \label{fig:calibration_gap}
\end{figure}

Figure~\ref{fig:calibration_gap} shows the calibration gap broken out by
question type.
Gemini 3.1 Flash-Lite exceeds the random Brier baseline in every category.
Severe overconfidence on spatial and word-lookup questions drives most
of the miscalibration, while even that model is slightly underconfident on
unknowable questions.
The best-calibrated models and the human tester show modest underconfidence
on unknowable questions, which is preferable to overconfidence in most deployment
scenarios.

\subsection{Refusal Rates}

A refusal occurs when a model produces a response that cannot be parsed as a valid
A--D answer, including content-policy refusals, uncertainty disclaimers, and
free-text answers without a parseable choice.
Refused responses are scored at $25\%$ confidence as described in
Section~\ref{sec:design}.
Word-lookup questions account for the majority of refusals, concentrated in
the OpenAI and Gemini 2.5 families.
The majority of these word-lookup refusals were copyright-related, where models
declined to reproduce text from copyrighted works.
Anthropic and Gemini 3.x models show near-zero refusals.
These models express uncertainty through lower stated confidence rather than
declining to answer.
Refusal rates per model are shown in Appendix~\ref{app:additional_figs}.

\subsection{Confidence Distributions}

\begin{figure}[htbp]
  \centering
  \includegraphics[width=0.48\textwidth]{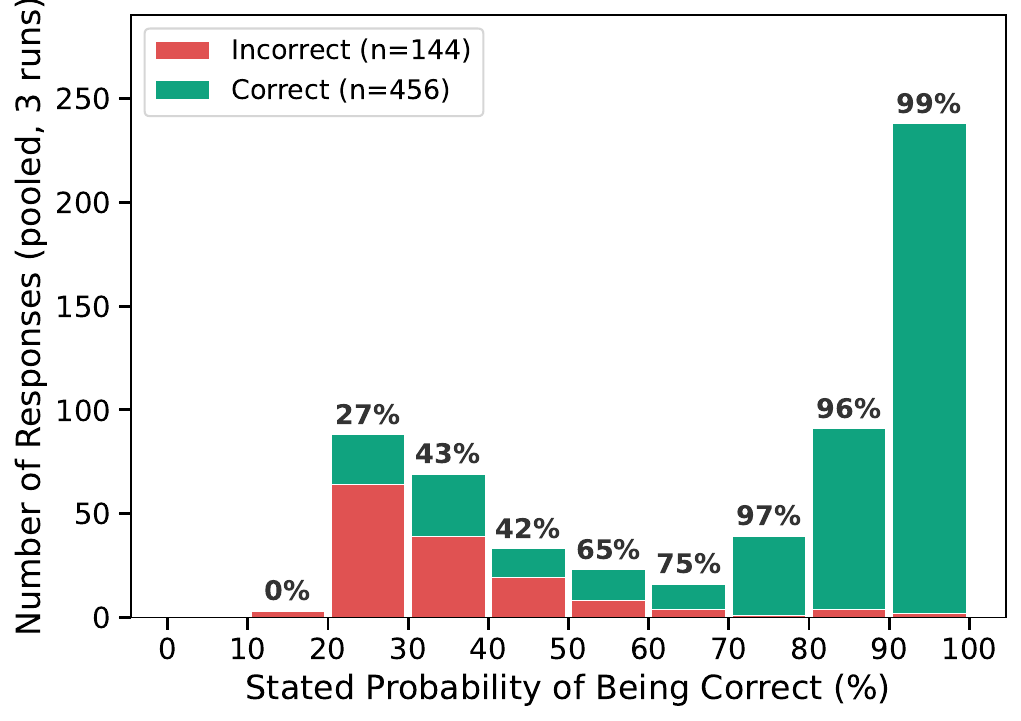}
  \hfill
  \includegraphics[width=0.48\textwidth]{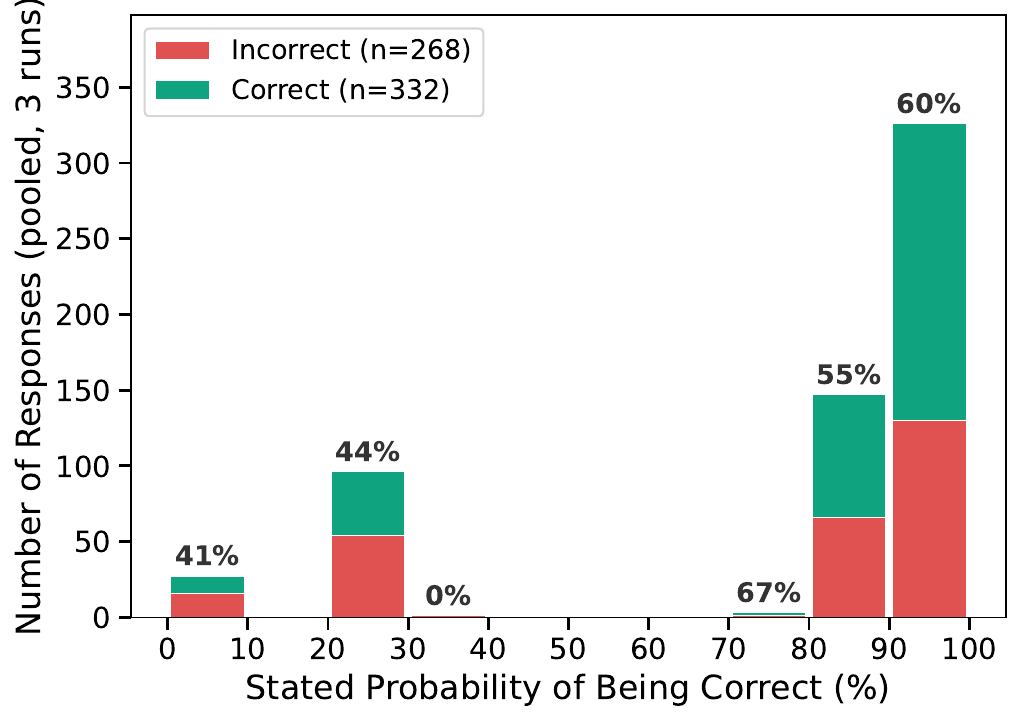}
  \caption{Stated probability distributions for Claude Opus 4.6 (left;
  Brier $0.103$, best-calibrated) and Gemini 3.1 Flash-Lite (right;
  Brier $0.367$, worst-calibrated).
  Numbers above bars: percentage of responses in that bin that were correct.
  Claude Opus 4.6 produces a spread distribution with clear monotone separation
  between correct and incorrect responses.
  Gemini 3.1 Flash-Lite shows a bimodal pattern: responses cluster near $25\%$
  or in the $80$--$100\%$ range.}
  \label{fig:prob_dist}
\end{figure}

Figure~\ref{fig:prob_dist} shows stated probability distributions for the best
and worst calibrated models.
Claude Opus 4.6 produces a spread distribution spanning the full
confidence range, with clear monotone separation between correct and incorrect
responses.
Gemini 3.1 Flash-Lite, by contrast, shows a bimodal pattern: nearly all
responses fall either near $25\%$ (the ``I don't know'' floor) or in the
$80$--$100\%$ range.

\subsection{Key Findings}

\paragraph{Reasoning effort and model scale both predict calibration quality.}
Within the GPT-5 family, reasoning effort broadly tracks calibration: Standard
($0.117$) and High ($0.121$) outperform Low ($0.141$), and higher reasoning also
tends to reduce run-to-run instability.
More broadly, the larger model within a generation consistently achieves a lower
Brier score across all families: Claude Opus variants outperform Claude Sonnet
variants at each release, Gemini 3.1 Pro Preview ($0.103$) substantially
outperforms Gemini 3.1 Flash-Lite ($0.367$), and Gemini 2.5 Pro outperforms
Gemini 2.5 Flash.
Reasoning compute and model scale thus appear to contribute independently to
calibration quality.

\paragraph{Newer models are not always better calibrated.}
Gemini 3.1 Flash-Lite is the worst-calibrated model in the study by a
wide margin (Brier $0.367$), despite being one of the most recently released
Google models, performing worse than models from the 2.5 generation.

\paragraph{High-performing models are underconfident on unknowable questions.}
On unknowable questions, models default to the $25\%$ ``I don't know''
floor even when they empirically guess correctly far more often than one-in-four.
This underconfidence is preferable to overconfidence for most deployment
scenarios and is visible in the calibration gap analysis
(Figure~\ref{fig:calibration_gap}).

\section{Discussion and Conclusion}

\begin{changedWblock}
ConfidenceBench evaluates verbalized confidence calibration in frontier LLMs using
a compact private benchmark and the Brier score as a proper scoring rule.
Across 15 models and 200 questions, we find that calibration is not captured by
accuracy alone: the most accurate model is not the best-calibrated, and several
models perform worse than a calibrated-random baseline despite reasonable accuracy.

Two models achieve lower Brier scores than the human tester, suggesting that
frontier LLMs can match or exceed the human baseline on targeted
multiple-choice calibration tasks.
Reasoning effort and model scale generally improve calibration, but newer models
are not always better calibrated.
These results suggest that confidence calibration should be evaluated directly,
especially in applications involving routing, abstention, or human-in-the-loop
oversight.
\end{changedWblock}

\paragraph{Limitations.}
\begin{changedWblock}
ConfidenceBench evaluates verbalized confidence elicited through prompting, rather
than confidence derived from model logits or internal probability distributions.
Reported probabilities may therefore partly reflect instruction-following behaviour
or prompt framing rather than the model's underlying epistemic uncertainty.

The benchmark is intentionally compact and targeted: it contains 200
English-language multiple-choice questions across four categories, so results may
not generalise to other domains, languages, cultural contexts, long-form generation,
or multi-turn interaction.
The question set is also kept private to reduce training contamination and preserve
future validity, which limits independent auditing of individual questions, although
the elicitation protocol, scoring rules, model list, and aggregate results are fully
reported.
\end{changedWblock}

\paragraph{Future work.}
\begin{changedWblock}
A primary next step is to maintain a public leaderboard tracking calibration scores
for newly released models.
Further extensions include comparing verbalized confidence with logprob-based
calibration, evaluating open-source and multimodal models, and expanding the
question set to additional domains.
\end{changedWblock}

\FloatBarrier

\section*{Broader Impacts}

\begin{changedWblock}
Better-calibrated confidence estimates may improve the reliability of LLM-based
systems by enabling abstention, escalation, and human review when model uncertainty
is high.
A potential misuse is that calibration benchmarks could help identify regimes where
models make high-confidence errors.
Keeping the question set private and reporting aggregate results reduces the risk of
benchmark-specific exploitation.
\end{changedWblock}

\bibliographystyle{abbrvnat}
\bibliography{refs}

\clearpage
\appendix

\section{Full Results and Run-to-Run Stability}
\label{app:full_results}

\begin{table*}[h]
\centering
\small
\caption{\changedW{Full results sorted by Brier score. $n$ = pooled non-refused
responses across 3 runs. Mean and standard deviation computed across 3 independent
runs; human tester evaluated once (no run-to-run variance). Accuracy in $[0,1]$.
Best Brier (Claude Opus 4.6, first by unrounded score), best Accuracy (Gemini 3.1
Pro Preview), and best ECE (Human Tester) bolded.}}
\label{tab:full_results}
\begin{tabular}{lrrrr}
\toprule
\textbf{Model} & $n$ & \textbf{Brier} $\downarrow$ & \textbf{Accuracy} $\uparrow$ & \textbf{ECE} $\downarrow$ \\
\midrule
Claude Opus 4.6         &  600 & $\mathbf{0.103} \pm 0.006$ & $0.760 \pm 0.018$ & $0.069 \pm 0.009$ \\
Gemini 3.1 Pro Preview  &  600 & $0.103 \pm 0.012$ & $\mathbf{0.827} \pm 0.008$ & $0.086 \pm 0.019$ \\
\midrule
Human Tester            &  200 & $0.105$\hphantom{${\pm}0.000$} & $0.705$\hphantom{${\pm}0.000$} & $\mathbf{0.053}$\hphantom{${\pm}0.000$} \\
\midrule
GPT-5 Standard          &  577 & $0.117 \pm 0.006$ & $0.766 \pm 0.022$ & $0.067 \pm 0.010$ \\
GPT-5 High              &  587 & $0.121 \pm 0.010$ & $0.758 \pm 0.005$ & $0.076 \pm 0.019$ \\
GPT-5 Mini              &  527 & $0.131 \pm 0.011$ & $0.721 \pm 0.032$ & $0.091 \pm 0.007$ \\
GPT-5 Nano              &  488 & $0.133 \pm 0.021$ & $0.705 \pm 0.025$ & $0.083 \pm 0.023$ \\
GPT-5 Low               &  581 & $0.141 \pm 0.016$ & $0.749 \pm 0.017$ & $0.078 \pm 0.027$ \\
Claude Sonnet 4.6       &  600 & $0.145 \pm 0.001$ & $0.690 \pm 0.010$ & $0.096 \pm 0.013$ \\
Claude Opus 4.5         &  600 & $0.149 \pm 0.006$ & $0.745 \pm 0.0$ & $0.141 \pm 0.002$ \\
Claude Sonnet 4.5       &  600 & $0.183 \pm 0.003$ & $0.708 \pm 0.006$ & $0.110 \pm 0.007$ \\
Claude Opus 4.1         &  600 & $0.191 \pm 0.005$ & $0.673 \pm 0.008$ & $0.124 \pm 0.013$ \\
Gemini 2.5 Pro          &  591 & $0.213 \pm 0.011$ & $0.694 \pm 0.011$ & $0.198 \pm 0.010$ \\
Gemini 2.5 Flash        &  561 & $0.227 \pm 0.005$ & $0.606 \pm 0.013$ & $0.188 \pm 0.008$ \\
Claude Sonnet 4         &  600 & $0.251 \pm 0.002$ & $0.578 \pm 0.003$ & $0.169 \pm 0.003$ \\
Gemini 3.1 Flash-Lite   &  600 & $0.367 \pm 0.017$ & $0.553 \pm 0.003$ & $0.344 \pm 0.023$ \\
\bottomrule
\end{tabular}
\end{table*}

\clearpage
\section{Reliability Diagrams}
\label{app:reliability}

\begin{figure}[H]
  \centering
  \includegraphics[width=0.75\textwidth]{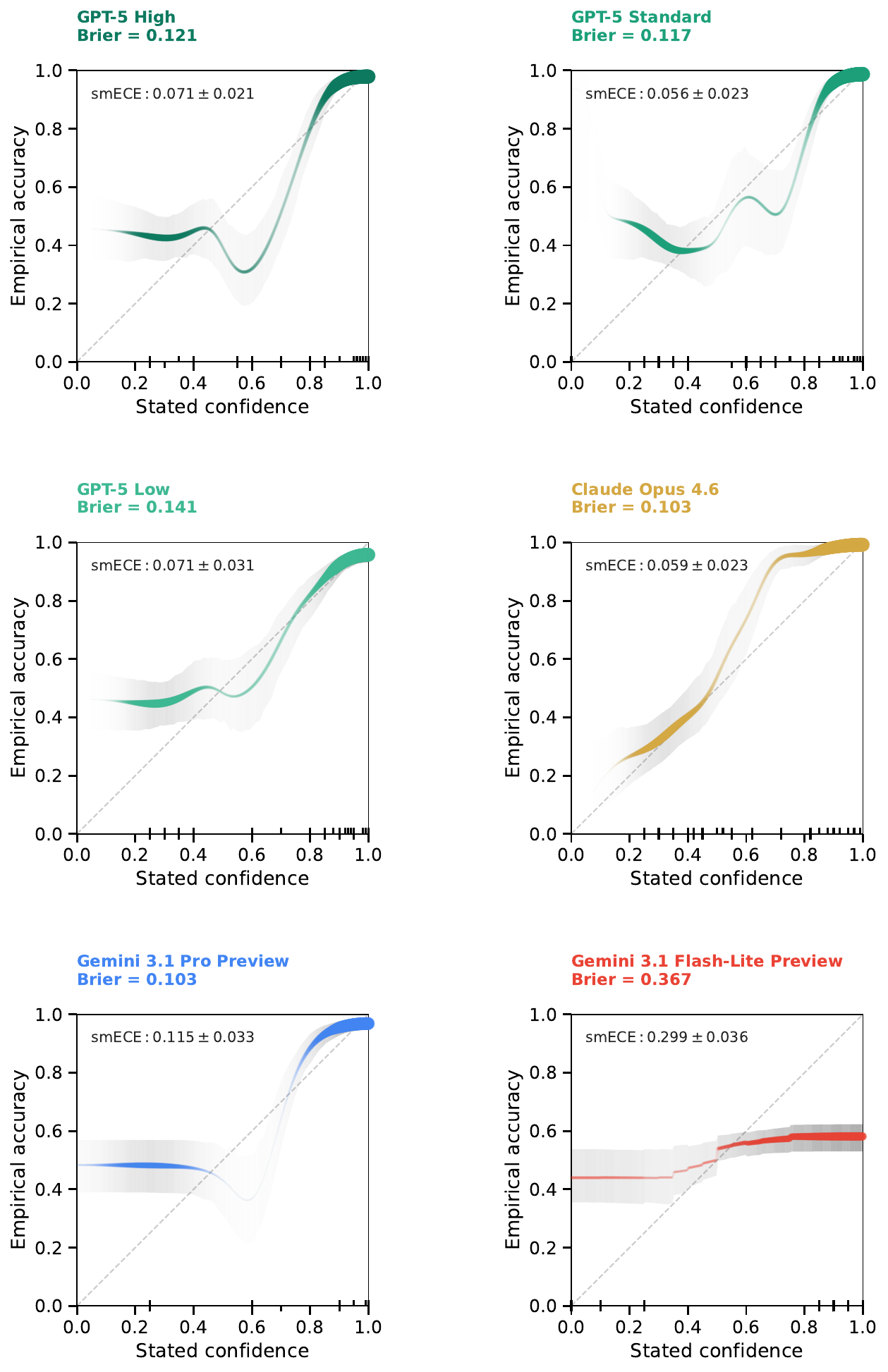}
  \caption{Kernel-smoothed reliability diagrams \citep{blasiok2023smoothece}.
  Dashed diagonal = perfect calibration.
  \changedW{Claude Opus 4.6 tracks the diagonal closely (smECE $0.059$);
  Gemini 3.1 Pro Preview shows moderately higher calibration error
  (smECE $0.115$) despite its strong Brier score.
  Gemini 3.1 Flash-Lite shows a near-flat curve with little relationship
  between stated confidence and empirical accuracy.}}
  \label{fig:reliability}
\end{figure}

\clearpage
\section{Elicitation Prompt Template}
\label{app:prompt}

The following prompt suffix is appended to each question (after the question text
and four labelled options):

\begin{quote}
\textit{You must also give a probability that your answer is correct, as a whole
number from 0 to 100. For example, 100 means you are completely certain, 50 means
you think it is a coin flip, and 25 means you are guessing randomly among the 4
options. Please provide your answer as a valid JSON object with the keys `answer'
and `probability', where `answer' is a single letter (A, B, C, or D) and
`probability' is a whole number from 0 to 100.
Example: \{"answer": "A", "probability": 75\}}
\end{quote}

\section{Additional Figures}
\label{app:additional_figs}

\begin{figure}[H]
  \centering
  \includegraphics[width=0.75\textwidth]{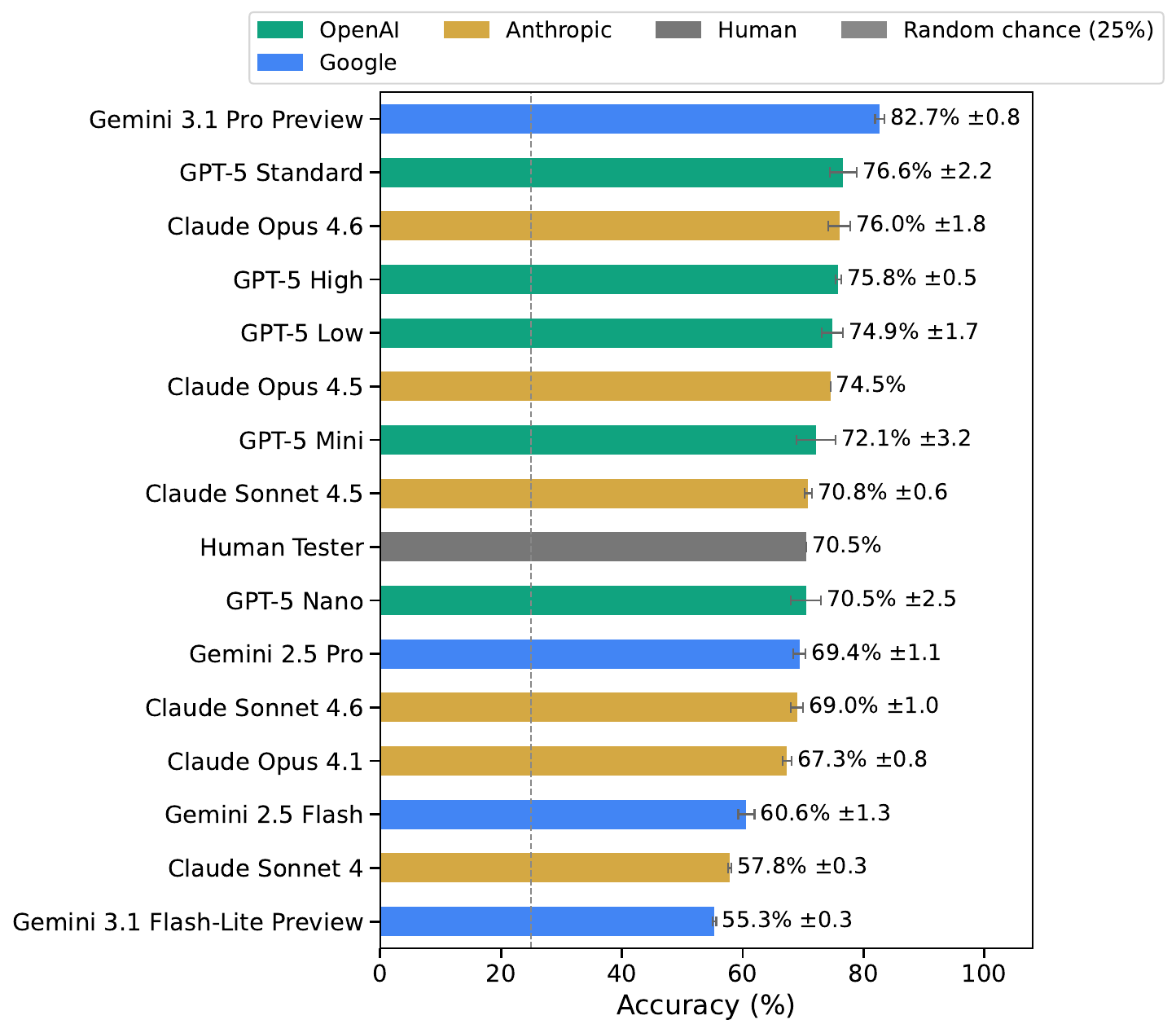}
  \caption{Accuracy by model (3-run means, error bars show std). Gemini 3.1 Pro
  Preview leads on accuracy ($82.7\%$), but Claude Opus 4.6 leads on Brier
  ($0.103$). Calibration and accuracy are not equivalent measures of model quality.}
  \label{fig:accuracy_by_model}
\end{figure}

\begin{figure}[H]
  \centering
  \includegraphics[width=0.75\textwidth]{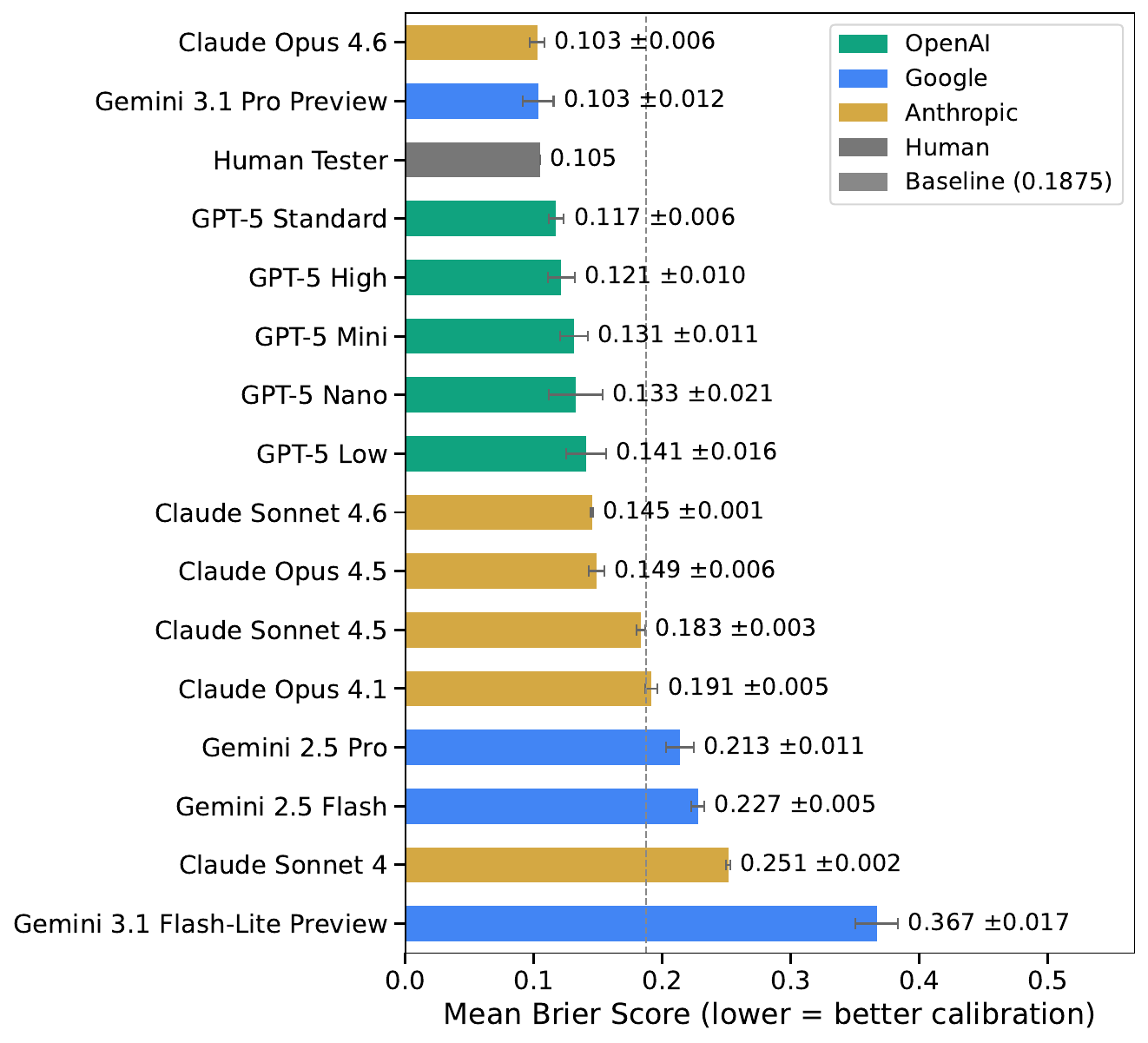}
  \caption{Mean Brier score by model (3 runs, 200 questions per model). Error
  bars show standard deviation across runs. Dashed line: calibrated-random
  baseline ($0.1875$). Human tester shown for reference.}
  \label{fig:brier_scores}
\end{figure}

\begin{figure}[H]
  \centering
  \includegraphics[width=0.75\textwidth]{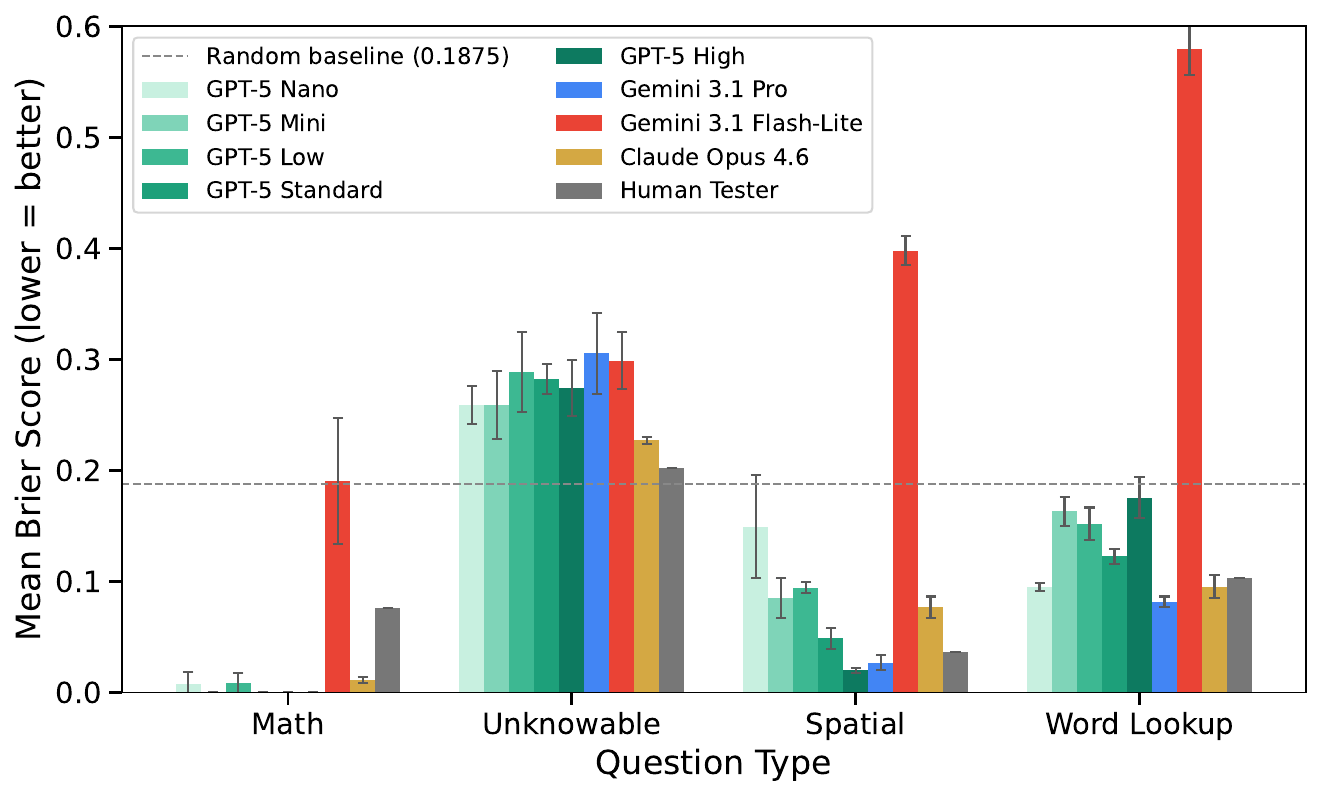}
  \caption{Mean Brier score by question type for selected models (3 runs, means).
  Dashed line: calibrated-random baseline ($0.1875$).
  Gemini 3.1 Flash-Lite exceeds the random baseline in every category.}
  \label{fig:brier_by_type}
\end{figure}

\begin{figure}[H]
  \centering
  \includegraphics[width=0.75\textwidth]{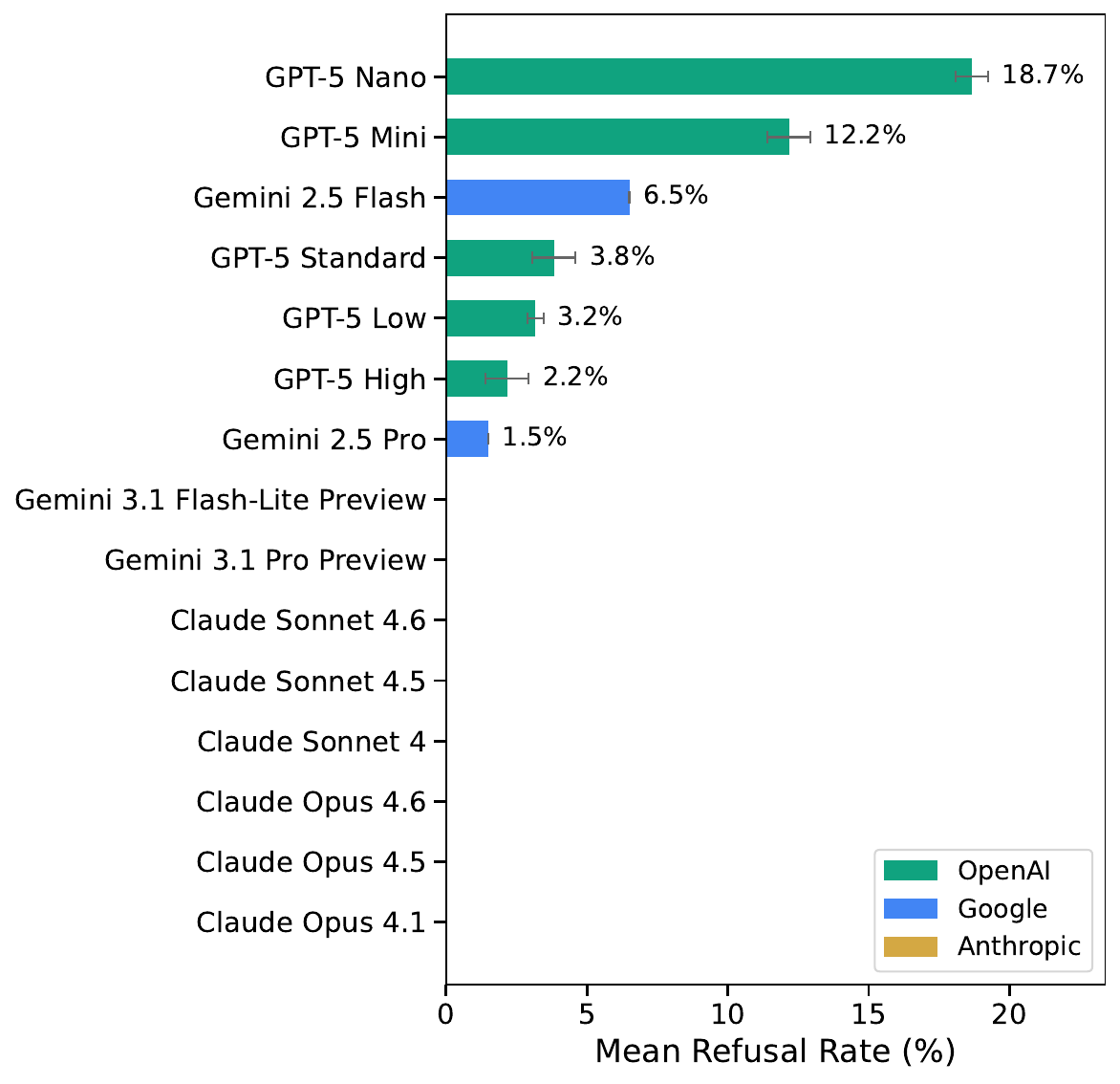}
  \caption{Total refusal rate per model (3-run means). Word-lookup refusals
  dominate across OpenAI and Gemini 2.5 families.
  GPT-5 Nano has the highest overall rate ($18.7\%$).}
  \label{fig:refusal_rate}
\end{figure}

Figures~\ref{fig:accuracy_by_model}, \ref{fig:brier_scores},
\ref{fig:brier_by_type}, and~\ref{fig:refusal_rate} provide supplementary views
that complement the main results.

\

\end{document}